\documentclass[preprint,12pt]{elsarticle}

\usepackage[utf8]{inputenc} 
\usepackage[T1]{fontenc}    
\usepackage{hyperref}       
\usepackage{url}            
\usepackage{booktabs}       
\usepackage{amsfonts}       
\usepackage{nicefrac}       
\usepackage{microtype}      
\usepackage{xcolor}         
\usepackage{subcaption}
\usepackage{graphicx}
\usepackage{eqnarray,amsmath}
\usepackage{multirow}
\usepackage{comment}
\usepackage{setspace}

\journal{a journal}

\begin{document}

\begin{frontmatter}

\title{\textit{FaceMixup}: Enhancing Facial Expression  Recognition through Mixed Face Regularization}

\author{$^1$Fabio A. Faria, $^1$Mateus M. Souza, $^2$Raoni F. da S. Teixeira, $^3$Maur\'icio P. Segundo}

\address{$^1$ Universidade Federal de S\~{a}o Paulo (Unifesp)\\
S\~ao Jos\'e dos Campos, SP -- Brazil \\
ffaria@unifesp.br
}

\address{$^2$Universidade Federal de Mato Grosso (UFMT)\\ 
Cuiab\'a, MT -- Brazil\\
}

\address{$^3$University of South Florida (USF)\\
Tampa, FL -- USA\\
}
\doublespacing
\begin{abstract}
The proliferation of deep learning solutions and the scarcity of large annotated datasets pose significant challenges in real-world applications. Various strategies have been explored to overcome this challenge, with data augmentation (DA) approaches emerging as prominent solutions. DA approaches involve generating additional examples by transforming existing labeled data, thereby enriching the dataset and helping deep learning models achieve improved generalization without succumbing to overfitting. In real applications, where solutions based on deep learning are widely used, there is facial expression recognition (FER), which plays an essential role in human communication, improving a range of knowledge areas (e.g., medicine, security, and marketing). In this paper, we propose a simple and comprehensive face data augmentation approach based on mixed face component regularization that outperforms the classical DA approaches from the literature, including the MixAugment which is a specific approach for the target task in two well-known FER datasets existing in the literature.
\end{abstract}


\end{frontmatter}

\doublespacing
\section{Introduction}
\label{sec:intro}

Deep convolutional neural network (CNN) has provided major advancements for computer vision and machine learning areas in many tasks such as segmentation~\cite{segmentation_2022}, object recognition~\cite{object_2023}, biometric recognition~\cite{biometric_2023} and medical image analysis~\cite{zhou2017deep}. When aiming to improve effectiveness in the target task, some ways emerge, such as data augmentation approaches (\emph{e.g.}, Cutout~\cite{cutout_2017}, AutoAugment~\cite{aa_cvpr2019}, and Mixup~\cite{mixup_iclr2018}), regularization approaches~\cite{regularization_2022} (\emph{e.g.}, Dropout~\cite{dropout_2014}, Label Smoothing~\cite{labelsmoothing_2019}, and Batch Augment~\cite{Augmentbatch_2019}) and training strategy (\emph{e.g.}, Curriculum learning~\cite {cvlearning_2022}).

Data augmentation (DA) approaches are particularly popular in computer vision tasks and play a crucial role in deep learning-based solutions for real applications, mainly due to the lack of large collections of labeled data. Therefore, DA allows increasing the dataset (volume, quality, and diversity), improving generalization model, helping in regularization, and domain adaptation~\cite{Mumuni2022}. Furthermore, DA can be applied in other domains, such as natural language processing~\cite{aiopen_2022,NLP_DA_NEURIPS2023} and audio processing~\cite{mva_2022}. 

Traditional geometric (\emph{e.g.}, random affine transformations) and photometric (\emph{e.g.}, color and contrast perturbation) image augmentation approaches knowingly perform consistently over a wide variety of computer vision solutions built with deep learning~\cite{Mumuni2022}. Over the past decade, additional general image augmentations have been presented~\cite{cutout_2017,mixup_iclr2018,cutmix_cvpr2019,Zhong2020}, some of them benefiting facial analyses just as well~\cite{Das2021,Psaroudakis2022_mixaugment,Borges2021}. In addition, other works presented face-specific augmentations to take advantage of experts' insights during training~\cite{Das2021,Lv2016,Psaroudakis2022_mixaugment,wang_NCA2020}. 

Face data augmentation is fundamentally important for improving the performance of neural networks in the following aspects. (1) It is inexpensive to generate a huge number of synthetic data with annotations in comparison to collecting and labeling real data. (2) Synthetic data can be accurate, so it has groundtruth by nature. (3) If controllable generation method is adopted, faces with specific features and attributes can be obtained. (4) Face data augmentation has some special advantages, such as generating faces without self-occlusion and balanced dataset with more intra-class variations~\cite{wang_NCA2020}.

Facial Expression Recognition (FER) task has benefited from face data augmentation approaches, which focus on discerning and classifying emotional expressions exhibited on a human face~\cite{Psaroudakis2022_mixaugment}. The objective is to streamline the process of identifying emotions by scrutinizing facial components like eyebrows, eyes, mouth, and other features, and correlating them with a range of emotions including anger, fear, surprise, sadness, and happiness. Furthermore, FER task plays an important role in a variety of real-world applications by the desire to enhance human-computer interaction, understand human emotions and behavior, improve marketing and advertising strategies, enhance security and surveillance systems, and facilitate applications in healthcare and assistive technologies~\cite{DFER_2022,survey_fer_2023}. 

In this paper, we propose a new simple and effective face data augmentation approach based on mixed facial regularization that creates new mixed faces by applying cropping and replacement operations between facial components of images of different classes and  taking these classes into consideration throughout the backpropagation of the deep neural network on the training process.

\section{Background \& Related Works}
\label{sec:relwork}


\subsection{General augmentations in facial analysis}
\label{sec:aug:general}

General augmentation methods pay no attention to facial semantics. Nevertheless, they can still be effective in face processing tasks.
\textbf{Cutout}~\cite{cutout_2017} builds upon dropout regularization~\cite{Srivastava2014_dropout} by discarding contiguous segments of an image instead of arbitrary pixel locations. Its implementation consisted of randomly masking square regions from input samples. This technique compels neural networks to take full advantage of the input image instead of focusing only on localized features. Figure~\ref{fig:da}(c) shows a face image after a Cutout operation. \textbf{Random Erasing}~\cite{Zhong2020} extends this concept to consider areas with different aspect ratios and to include random values in the excluded area, as shown in Figure~\ref{fig:da}(d). 
\textbf{Mixup}~\cite{mixup_iclr2018} performs a weighted sum of random image pairs from the training data and their respective one-hot labels. The mixing factor $\lambda\in[0,1]$ is also chosen randomly. This technique addresses problems related to memorization and sensitivity to adversarial examples by making decision boundaries smoother, which is accomplished by filling the area between classes in the problem space. An example of the Mixup of two face images is presented in Figure~\ref{fig:da}(e). 
\textbf{CutMix}~\cite{cutmix_cvpr2019} was later proposed as a solution to the lack of realism in images produced by Mixup. It consists of replacing a random rectangle of an image with the corresponding area from another image. One-hot labels are also combined, but in this case, they reflect the proportion of area from each source image. Figure~\ref{fig:da}(f) shows one example of CutMix for two face images. 
Lastly, instead of designing new augmentations, \textbf{AutoAugment}~\cite{aa_cvpr2019} seeks to select the ideal set of general augmentations for a certain problem from a fixed list with many pre-existing geometric and photometric approaches plus Cutout and SamplePairing, which is analogous to Mixup. 

The discussed augmentation operations successfully boosted the performance of face deepfake detection~\cite{Das2021}, facial expression recognition~\cite{Psaroudakis2022_mixaugment,Greco2023}, and face recognition~\cite{Borges2021}. Nevertheless, researchers found many opportunities to further improve these results by devising face-specific augmentation approaches, as discussed next.

\subsection{Face-specific augmentations}

We divide these approaches into learned~\cite{He2019_attgan,choi2018_stargan,Deng2020} and handcrafted~\cite{Psaroudakis2022_mixaugment,Das2021,Lv2016}. Both conform to face semantics, but while learned approaches use pretrained generative models to edit certain facial attributes, handcrafted approaches rely on researchers’ ingenuity to devise functional face modifiers. 

\subsubsection{Learned approaches} 

The evolution of generative models~\cite{Taylor2022_generative}, whose unprecedented level of realism led to their widespread use, found applications in face analysis too. For example, representing face variations adequately for all training subjects can be a challenging task when creating a dataset. But one could use generative models~\cite{He2019_attgan,choi2018_stargan,Deng2020} for  \textbf{attribute modification} (\emph{e.g.}, hairstyle, makeup, pose, expression, and age) to augment the number of samples per identity and enhance the models’ understanding of intra-class and inter-class boundaries. Figures~\ref{fig:da}(g)~and~\ref{fig:da}(h) show two examples where facial attributes -- hair color and expression -- from Figure~\ref{fig:da}(b) were modified using AttGAN~\cite{He2019_attgan}. This attribute modification strategy was successfully used to improve the performance of face recognition trained on synthetic data~\cite{Boutros2023} and expression recognition in speaking subjects~\cite{Song2019}. 

Among the advantages of learned approaches, we can mention their one-fits-all capacity (\emph{i.e.}, the same generative model can perform different augmentation operations) and ease of upgrade (\emph{i.e.}, replacing old models with newer, better ones). However, they can be computationally expensive, require substantial annotation labor for training, reproduce biases from training data, and modify attributes that should remain unchanged (\emph{e.g.}, identity traits modified while changing hair color in Figure~\ref{fig:da}(g)). Handcrafted approaches offer more control over the result, even though sometimes this means sacrificing photo-realism.

\subsubsection{Handcrafted approaches}

\begin{figure*}[ht!]
    \centering
    \includegraphics[width=1\linewidth]{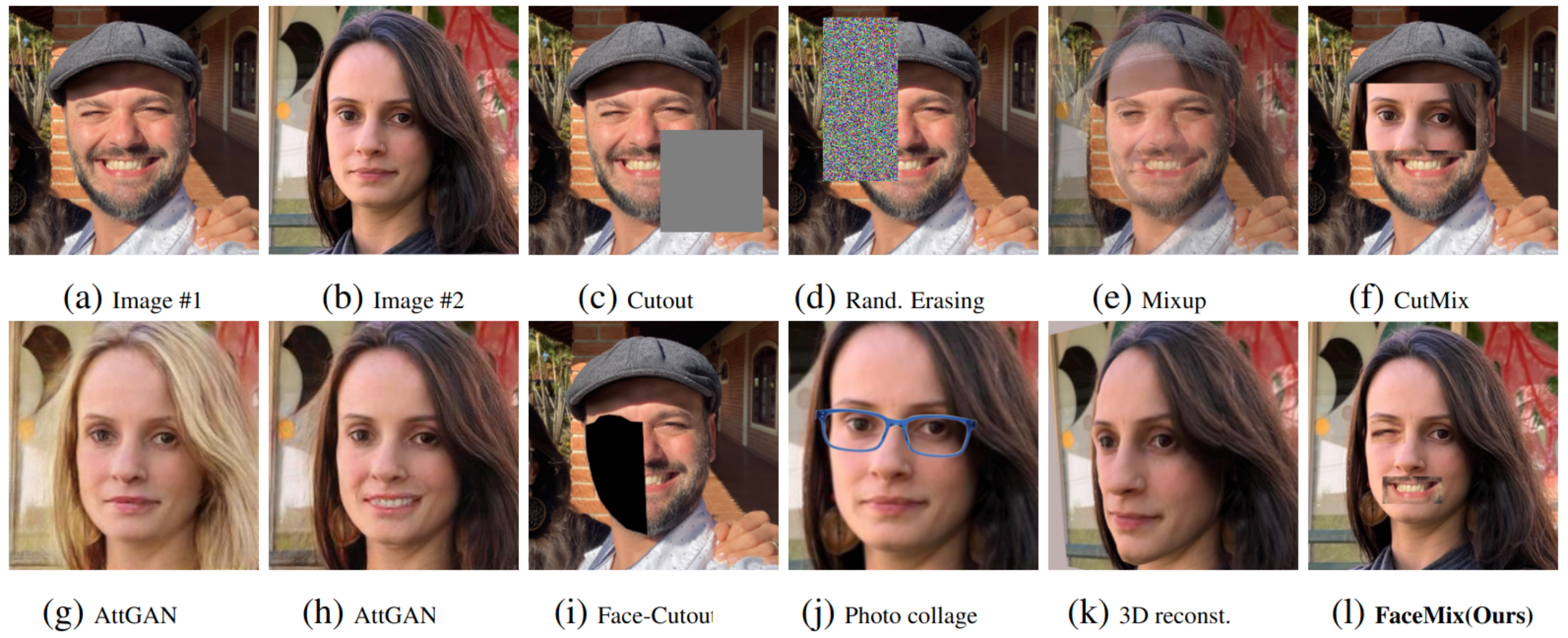}
    \caption{Visual comparison of different data augmentation approaches..}
    \label{fig:da}
\end{figure*}

Face recognition has been a popular research topic since the early 1990s, and multiple works addressed the normalization of the most common variations (\emph{\emph{e.g.},} pose, illumination, and accessories) before the deep learning era. As expected, this corpora inspired some augmentation operations. \textbf{Landmark perturbation}~\cite{Lv2016} seeks to increase robustness to face misalignment due to inaccurate landmark localization by injecting noise into the landmark coordinates used for image preprocessing during training. The outcome, however, is very similar to traditional geometric augmentation. \textbf{Photo collages}~\cite{Lv2017} were used to change hairstyle and accessories (Figure~\ref{fig:da}(j) illustrates the addition of eyeglasses), and \textbf{3D reconstruction}~\cite{Lv2017} enabled the creation of new poses and illumination conditions (Figure~\ref{fig:da}(k) illustrates a new pose). More recent works focused on adapting newer and simpler general augmentation approaches (see Section~\ref{sec:aug:general}) to a context of face analysis. \textbf{Face-Cutout}~\cite{Das2021} extends Cutout to use facial landmarks to define regions of the face to be masked out, such as eyes, nose, mouth, or even an entire side of the face, instead of masking random image regions. Figure~\ref{fig:da}(i) shows one example of this operation. \textbf{MixAugment}~\cite{Psaroudakis2022_mixaugment} updates Mixup to avoid issues when merging faces with different head poses. It combined mixed and unmodified images into the training loss to this end.

Face-specific handcrafted augmentation also improved the accuracy of face deepfake detection~\cite{Das2021}, facial expression recognition~\cite{Psaroudakis2022_mixaugment}, and face recognition~\cite{Lv2017,Lv2016}. Most of these approaches rely on facial landmarks, which can be time-consuming to obtain. But this information can be precomputed so that the augmentation can be performed without major computational effort during training.

Our work presents a novel face-specific handcrafted augmentation grounded on two observations. First, face-specific adaptions of general augmentation approaches (\emph{e.g.}, Face-Cutout and MixAugment) outperform their original counterparts (\emph{e.g.}, Cutout and Mixup) in facial analyses~\cite{Psaroudakis2022_mixaugment,Das2021}. Second, CutMix outperforms Mixup and Cutout as a general augmentation approach~\cite{cutmix_cvpr2019}. With that in mind, we proposed \textbf{\textit{FaceMixup}}, an adaptation of CutMix that mixes facial components (\emph{e.g.}, eyebrows, eyes, nose, and mouth) from image pairs, but it also differs from CutMix due to the existence of semantic information in the process of replacing the region of interest. Furthermore, when in uncontrolled environment, an extended version (\textbf{\textit{FaceMixup}$+$RS}) incorporates mixed and unmodified images into the training process, as in MixAugment. Details on its operation are presented in the following section.

\section{\textit{FaceMixup} Approach}
\label{sec:proposal}
In this section, the two stages of the \textit{FaceMixup} approach are described in detail.

\subsection{Creation of the Mixed Face Dataset}

First, given the training set from an original face dataset, for each the pairs of face images $(\mathcal{I}_i$ and $\mathcal{I}_j$) with their respectively classes ($\mathcal{Y}_i$ and  $\mathcal{Y}_j$) are sampled with the aim of being combined in the final process.
Let $\mathcal{I}_i$ be the ``supplier" face image, which provides the facial components to be pasted into the ``receiver" face image $\mathcal{I}_j$. Next, the $\gamma$ parameter is chosen among six possible values $\gamma=\{1,2,3,4,5,6\}$, which it means that each new mixed face $\hat{{\mathcal{T}}}$ can have up to six face components replaced at the same time.

The choice of which $\gamma$ facial components (\emph{e.g.}, 1: Left Eyebrow -- EB, 2: Right Eyebrow -- RB, 3: Left Eye -- LE, 4: Right Eye -- RE, 5: Nose -- NO, and 6: Mouth -- MO) will be cut from $\mathcal{I}_i$ and paste on the $\mathcal{I}_j$ is also a random decision. Therefore, once all six facial components of both face images (``supplier" and ``receiver") are detected by DLib algorithm~\cite{dlib09}, cropping $\mathcal{I}_i$ and replacing the $\gamma$ facial components in the face image $\mathcal{I}_j$ are performed in sequence. 

\begin{figure*}[ht!]
    \centering
    \includegraphics[width=1\linewidth]{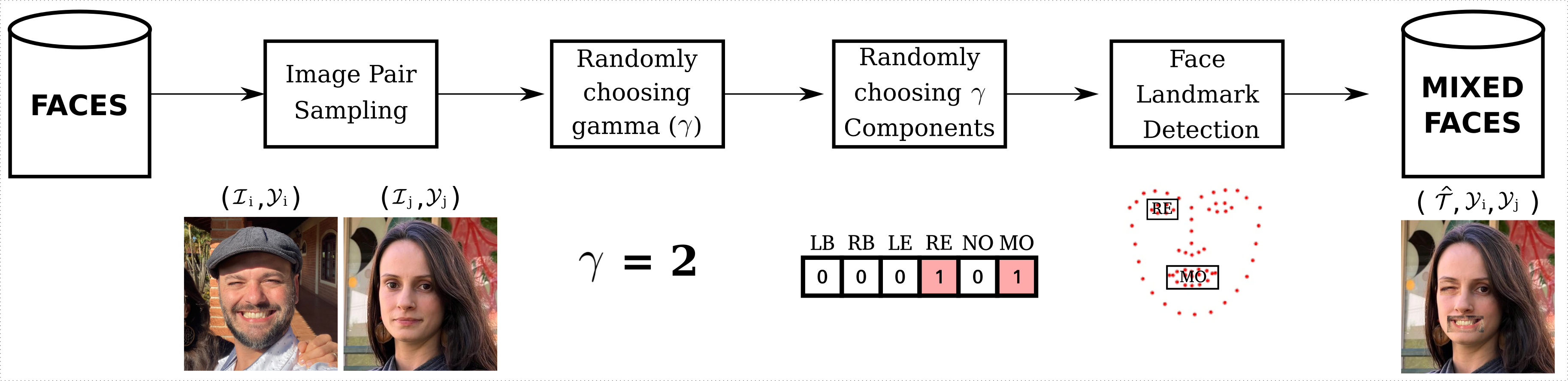}
    \caption{ The creation process of the mixed face dataset proposed in this work.}
    \label{fig:mixedface}
\end{figure*}

In this sense, it is very important to observe that in this process of creation of the mixed face dataset, given a dataset composed of $N$ face images, it can create $P_{N,2} \times (C_{6,1}+C_{6,2}+C_{6,3}+C_{6,4}+C_{6,5}+C_{6,6})=P_{N,2} \times 62 = \frac{N!}{(N-2)!} \times 62 $
possible new mixed face images $\hat{{\mathcal{T}}}$, where $P_{N,2}$ is the permutation of $N$ face images and the value $2$ means the pair of faces (``supplier" and ``receiver").  Figure~\ref{fig:mixedface} presents the creation process of the mixed face dataset. For example, a dataset consisting of $1,000$ face images can create more than $61$M new mixed face images.

\subsection{Mixed Face Regularization Approach}
Once the mixed face dataset has been created, a deep neural network architecture is learned for the target task.
Let $CE(\mathcal{I}_i,\mathcal{Y}_{i}) = p(\mathcal{I}_i)\times \log p(\mathcal{Y}_i)$ be the cross-entropy function for a given image $\mathcal{I}_i\in{\cal I}$, where $p(\mathcal{I}_i)$ denotes the probability vectors extracted from the output of the CNNs and $p(\mathcal{Y}_i)$ denotes its corresponding label vector (one-hot vector). The \textit{FaceMixup} loss function is computed as follows:
\begin{equation}
  {\cal L}_{FaceMixup} =  \frac{\gamma}{W} \times CE(\hat{{\mathcal{T}}},{\mathcal{Y}_i}) + (1-\frac{\gamma}{W}) \times CE(\hat{{\mathcal{T}}},{\mathcal{Y}_j}), 
  \label{Eq:mixup}
\end{equation}

\noindent where $\mathcal{Y}_i\neq \mathcal{Y}_j$, $\gamma=\{1,2,3,4,5,6\}$ corresponds to the number of face components to be replaced, and $W$ is a weighting factor for the mixed face image  $\hat{{\mathcal{T}}}$. This $W$ value is defined during the  experimental stage.

For in-the-wild face image dataset, an extended version of the \textit{FaceMixup} approach called ${\cal L}_{FaceMix+RS}$ has been proposed in this work. In this loss function version, the crossentropy function under the real samples is added to complete the final loss function.

\begin{eqnarray}
{\cal L}_{RS}  = CE({\mathcal{I}_{i}},\mathcal{Y}_{i}) + CE(\mathcal{I}_{j},\mathcal{Y}_{j})\nonumber\\
{\cal L}_{FaceMix+RS} =  {\cal L}_{FaceMixup} + {\cal L}_{RS}, 
\label{Eq:FaceMix}
\end{eqnarray}

\begin{figure}[ht!]
    \centering
    \includegraphics[width=0.65\linewidth]{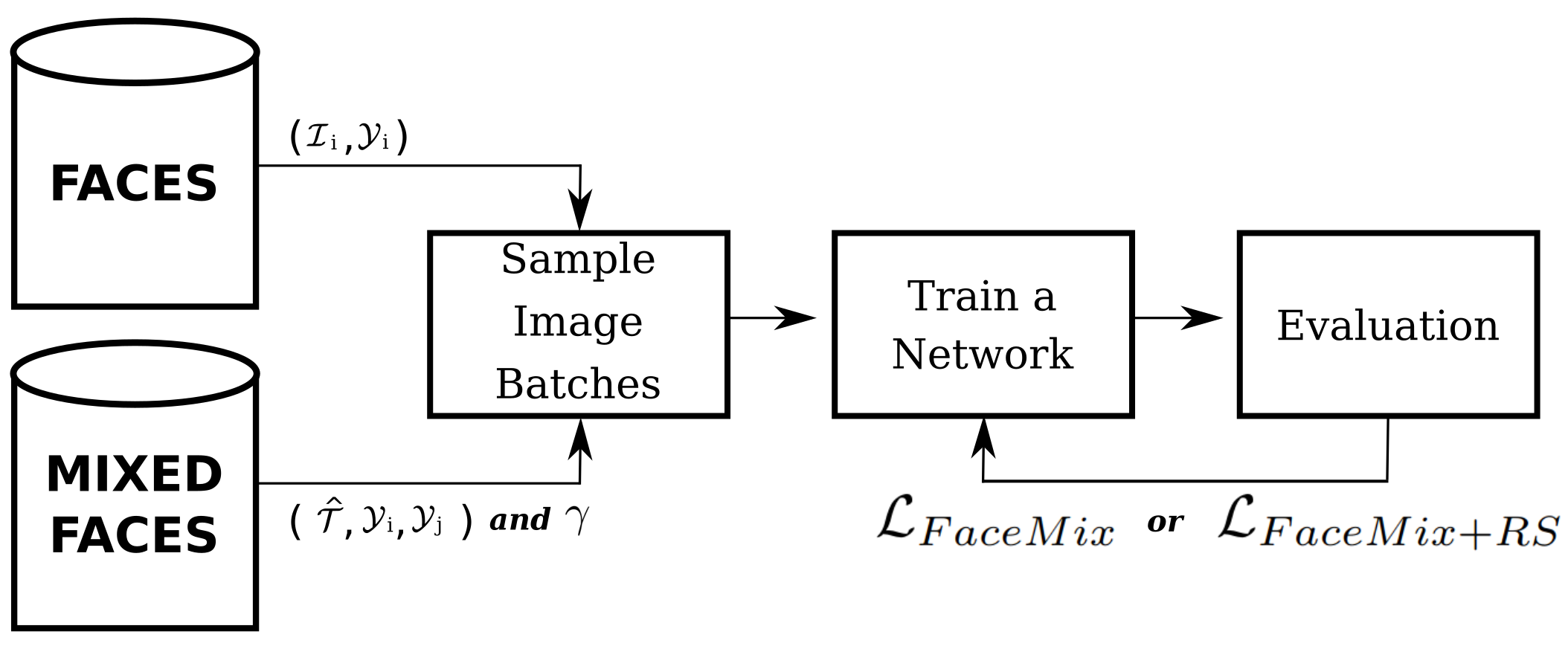}
    \caption{ The mixed face regularization approach for improving facial expression recognition task.}
    \label{fig:regula}
\end{figure}

Figure~\ref{fig:regula} illustrates the mixed face regularization approach during training process of the neural network.

\section{Experiments}
\label{sec:methodoly}
We experimentally investigate and compare the performance of \textit{FaceMixup}.

\subsection{Datasets}
\label{sec:datasets}
In this work, we have adopted two different image datasets. First, the FACES~\cite{faces} is a dataset composed of more controllable facial images. Next, we used a more challenging dataset, the AffectNet~\cite{affectnet}, which is composed of in-the-wild images (uncontrolled environment). Finally, it is very important to comment that it was not possible to adopt the same experimental protocol used in the MixAugment~\cite{Psaroudakis2022_mixaugment} paper because the Real World Affective Faces Database (RAF-DB) was not made available by its creators. 
For this reason, it was not possible to perform experiments with the RAF-DB dataset.


\subsubsection{FACES}
\label{sec:faces}
{FACES}\footnote{FACES dataset -- \url{https://faces.mpdl.mpg.de/imeji/}} is an image database of naturalistic faces of 171 young (n = 58), middle-aged (n = 56), and older (n = 57) women and men displaying each of six facial expressions: neutral, sad, disgust, fear, anger, and happy. The FACES database was developed between $2005$ and $2007$ by Natalie C. Ebner, Michaela Riediger, and Ulman Lindenberger at the Center for Lifespan Psychology, Max Planck Institute for Human Development, Berlin, Germany. The database comprises two sets of pictures per person and per facial expression ($a$ vs. $b$ set), resulting in a total of $2,052$ images with dimensions of $2835 \times 3543$.  

In this work, the FACES dataset has been separated into $1032$ images to training set and $1020$ images to test set, i.e., $86$ individuals have been randomly selected for training set and $85$ individuals for test set. Furthermore, the MIXED FACES dataset created by the proposed method is composed of $5678$ images with different $\gamma$ values, which will be used by \textit{FaceMixup} approach in conjunction with the original training set, totaling  $6710$ images. Due to the FACES dataset usage policy, samples of the  MIXED FACES dataset cannot be published in this paper. Please, see a fictitious example in Figure~\ref{fig:da}(l).

\begin{figure*}[ht!]
    \centering
    \includegraphics[width=1\linewidth]{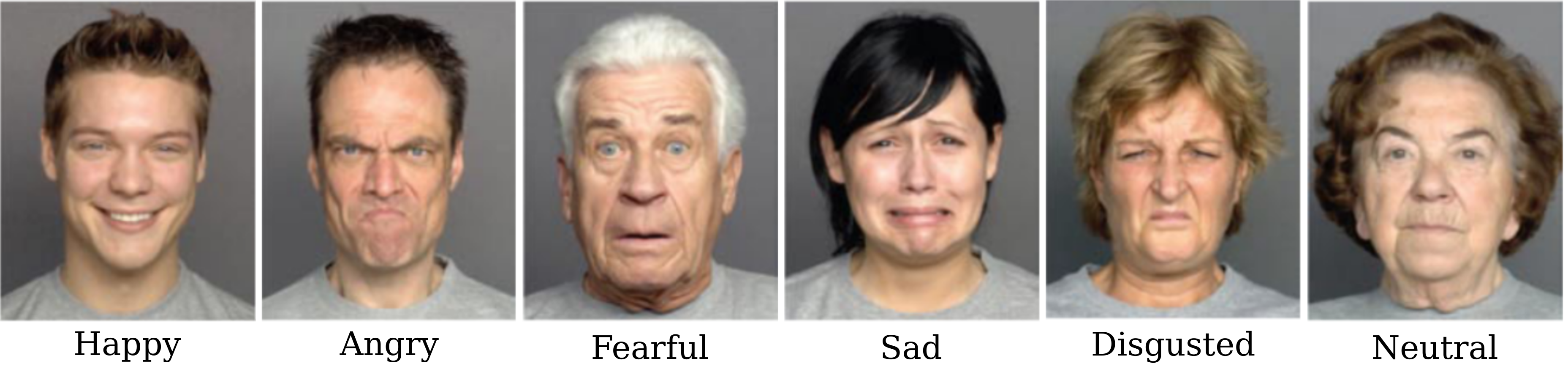}
    \caption{ Examples of six images from the FACES dataset, each with different facial expressions. Image adapted from~\cite{faces}.}
    \label{fig:faces_imgs}
\end{figure*}

\subsubsection{AffectNet}
\label{sec:affectnet}
AffectNet is one of the largest publicly available FER datasets, which  consists of two distinct benchmarks: AffectNet-7 and AffectNet-8. AffectNet-7 comprises 283,901 images for training and 3500 images for testing, with seven emotion categories including neutral (74,874), happy (134,415), sad (25,459), surprise (14,090), fear (6378), angry (24,882), and disgust (3803). AffectNet-8 introduces an additional category of ``contempt'' (3750) and expands the training set to 287,651 images, along with 4000 images for testing. 

\begin{figure*}[ht!]
    \centering
    \includegraphics[width=1\linewidth]{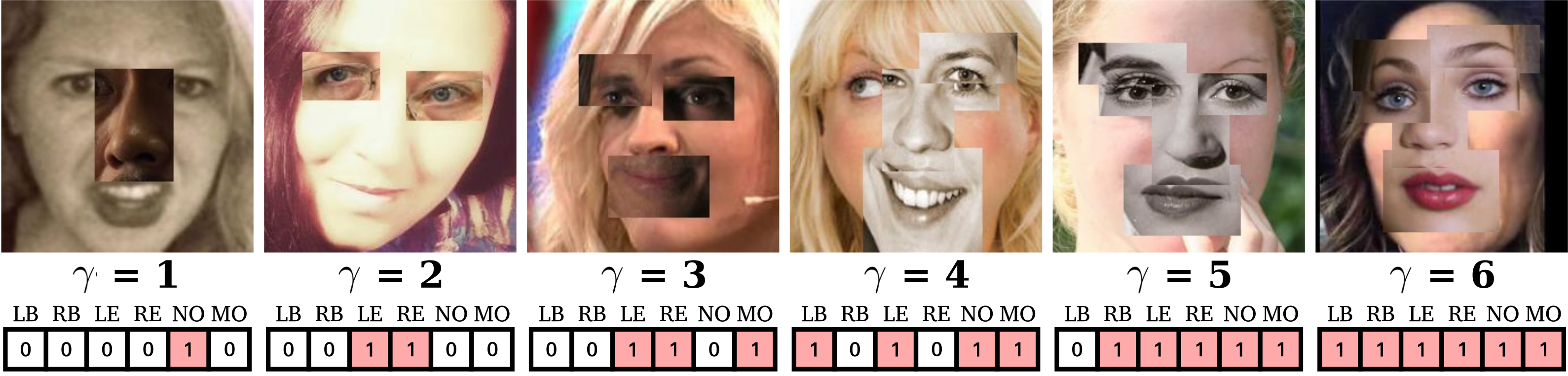}
    \caption{Examples of transformed images from the Mixed Faces dataset based on AffectNet dataset.}
    \label{fig:mixed_affectnet_imgs}
\end{figure*}

Due to the inability of high-processing resources, this work has used a subset of images randomly selected from the AffectNet-8 dataset ($1024$ images to training set and $1020$ images to test set) so that it would be possible to carry out experiments to validate the proposed approach on uncontrolled environment (in-the-wild scenario). Furthermore, the MIXED FACES dataset created by the proposed method is composed of  $5994$ images with different $\gamma$ values, which will be used by \textit{FaceMixup} approach in conjunction with the original training set, totaling $7018$ images. Figure~\ref{fig:mixed_affectnet_imgs} shows some examples of transformed images from the MIXED FACES dataset based on AffectNet dataset.

\begin{figure*}[ht!]
    \centering
    \includegraphics[width=1\linewidth]{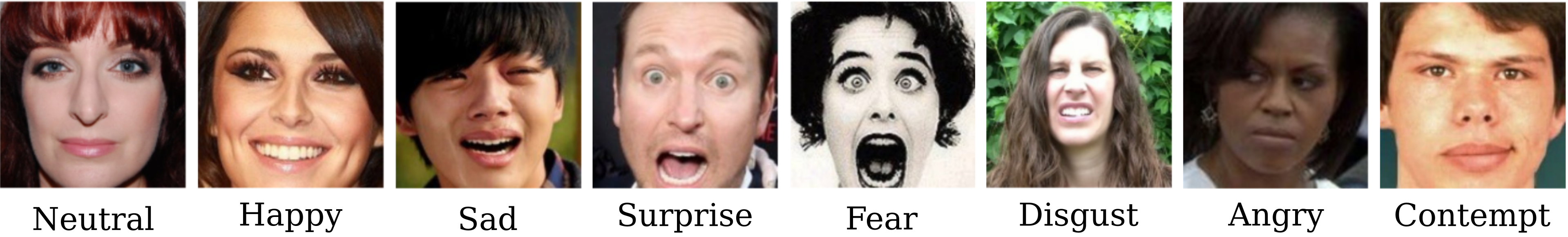}
    \caption{ Examples of in-the-wild images from the AffectNet dataset. Image adapted from~\cite{affectnet}.}
    \label{fig:affectnet_imgs}
\end{figure*}

\vspace{-0.25cm}

\subsection{Setup and Equipment}
In the performed experiments, the EfficientNet architecture (pre-trained EfficientNet-B0 from Pytorch Lib\footnote{pre-trained EfficientNet-B0 from Pytorch Lib -- \url{https://pytorch.org/hub/nvidia\_deeplearningexamples\_efficientnet/}}) used the following setup:  image dimensions of {$128 \times 128 \times 3$}, batch size$=64$, SGD optimizer (Weight decay $=1e-4$, Momentum{$=0.9$}, and Learning rate {$=3e-5$}), Dropout$=0.2$, and epochs$=200$. After performed experiments, the \textit{FaceMixup}-based approaches used parameter $W=7.9$ and $W=6.1$ for FACES and AffectNet datasets, respectively. Furthermore, $\gamma=\{1,2,3,4,5,6\}$ and six facial components such as Eyebrows (LB and RB), Eyes (LE and RE), Nose (NO), and Mouth (MO).

In general, the experiments were conducted on two workstations with the following configuration: (1) Intel Xeon E5-2630 v4 (40 cores) \@3.100GHz, 1x NVIDIA TITAN V and 64GB RAM; and (2) Intel Xeon E5-2630 v4 (40 cores) \@3.100GHz, 1x NVIDIA TITAN XP and 64GB RAM.

\subsection{Coarse-grained Analysis}

In the experiments carried out in this work, different approaches from the literature were compared with the approaches proposed (\textit{FaceMixup} and \textit{FaceMixup}$+$RS) on  two datasets of facial expression recognition tasks (FACES and AffectNet-8).

Table~\ref{tab:fer} shows the accuracy results of the Vanilla architecture (EfficientNet~\cite{efficientnet_pmlr2019} without Data Augmentation), Vanilla with Mixed Faces added to the training set, and four well-known data augmentation approaches in the literature (Mixup~\cite{mixup_iclr2018}, CutMix~\cite{cutmix_cvpr2019}, MixAugment~\cite{Psaroudakis2022_mixaugment} and Random Erasing~\cite{randomerasing_zhong2020random}). Furthermore, two of the most used approaches (AutoAugment -- AA~\cite{aa_cvpr2019} and Cutout~\cite{cutout_2017}) were combined with all other approaches compared in these experiments. 

In these performed experiments are two completely different scenarios. FACES is a dataset in a controlled environment (lab scenario) with all frontal faces filling the same region of the image, and following a designed template. Unlike the AffectNet-8 dataset, which the images are in uncontrolled and challenging environments (in-the-wild scenario), which it was not constructed or designed with research questions in mind, like FACES (See Section~\ref{sec:datasets}).

Regarding the accuracy results on FACES dataset, the Mixup-based approaches (Mixup, CutMix, and MixAugment) combined with AA present worse results compared to their original versions using EfficientNet architecture. However, the remaining approaches (Vanilla, Vanilla with Mixed Faces, Random Erasing, \textit{FaceMixup}, and \textit{FaceMixup}$+$RS) achieve better results when combined with AA. Almost the same behavior occurs when the approaches use of Cutout, except the MixAugment that achieve a small improvement. 

Unlike on accuracy results on the challenging AffectNet dataset, only Vanilla with Mixed Faces approach with Cutout achieve better results than it original version. With AA, only MixAugment, Random Erasing, \textit{FaceMixup}-based approaches have their accuracies increased.

Finally, as can be observed, the \textit{FaceMixup} and \textit{FaceMixup}$+$RS approaches achieved superior results than all baseline approaches on both datasets compared in this work. When the best baseline approach (MixAugment) is considered on FACES dataset, the \textit{FaceMixup}$+$RS approach achieved at least a relative gain of $2.46\%$. On the AffectNet dataset,  the \textit{FaceMixup}$+$RS approach achieved at least a relative gain of $7.78\%$ against the best baseline approaches (Vanilla with Mixed Faces and MixAugment). Furthermore, the relative gains are even greater if the version of the \textit{FaceMixup}$+$RS using AA is considered, with 5.44\% and 26.36\% for FACES and AffectNet datasets, respectively.

\vspace{-0.35cm}

\begin{table}[ht!]
    \centering
    \caption{Classification results of the data augmentation approaches  using EffecientNet (Eff.Net~\cite{efficientnet_pmlr2019}) for two different datasets (FACES and AffectNet). The colored highlights show the best results achieved by the DA approaches.}
    \resizebox{1\textwidth}{!}{
    \begin{tabular}{lccc|ccc} \hline
    \multirow{2}{*}{\textbf{Approaches}}&\multicolumn{3}{c|}{\textbf{FACES}}& \multicolumn{3}{c}{\textbf{AffectNet}} \\ \cline{2-7}
      & \textbf{Eff.Net}  &  \textbf{AA} &  \textbf{Cutout} & 
      \textbf{Eff.Net}  & \textbf{AA}& \textbf{Cutout} \\ \hline
    Vanilla & 80.78 & 84.51 & 83.63 & 24.11& 23.03&21.96 \\
    Vanilla {with Mixed Faces}  & 72.25 & 75.88 & 75.39 & \textcolor{red}{\textbf{27.35}} &27.05  &  \textcolor{red}{\textbf{28.92}}\\
    Mixup\cite{mixup_iclr2018}  & {82.16} & 66.37 & 74.02& 22.25 & 20.78 & 20.00\\
    CutMix\cite{cutmix_cvpr2019}  &  \multicolumn{1}{c}{83.92} & 81.47 & 82.15 &{22.54} & 21.47 & 20.88 \\   
    MixAugment\cite{Psaroudakis2022_mixaugment} & \textcolor{red}{\textbf{86.76}} &  \textcolor{red}{\textbf{86.56}} &  \textcolor{red}{\textbf{87.84}} & \textcolor{red}{\textbf{27.35}} & \textcolor{red}{\textbf{27.54}} & {26.07} \\
    Random Erasing\cite{randomerasing_zhong2020random} & 82.64 &  81.27 &  83.72& 20.88 & 22.35 & 23.52 \\
    \hline   
    \textit{FaceMixup}  \small{\textbf{(Ours)}} & {88.92} & {90.20} & {89.22} & {29.01} & {30.29} & {27.35} \\
    \textit{FaceMixup}  $+$ RS \small{\textbf{(Ours)}} & \textcolor{blue}{\textbf{88.92}} & \textcolor{blue}{\textbf{91.27}} & \textcolor{blue}{\textbf{90.00}} & \textcolor{blue}{\textbf{33.23}} & \textcolor{blue}{\textbf{34.80}} & \textcolor{blue}{\textbf{31.17}}  \\ \hline
    \textbf{Relative Gain}& \textbf{2.49\%} & \textbf{5.44\%}& \textbf{2.46\%} & \textbf{21.50\%} & \textbf{26.36\%}& \textbf{7.78\%} \\\hline
    \end{tabular} 
    }
    \label{tab:fer}
\end{table}

\subsection{Fine-grained Analysis} 
\begin{figure*}[ht!]
    \centering
    \includegraphics[width=1\linewidth]{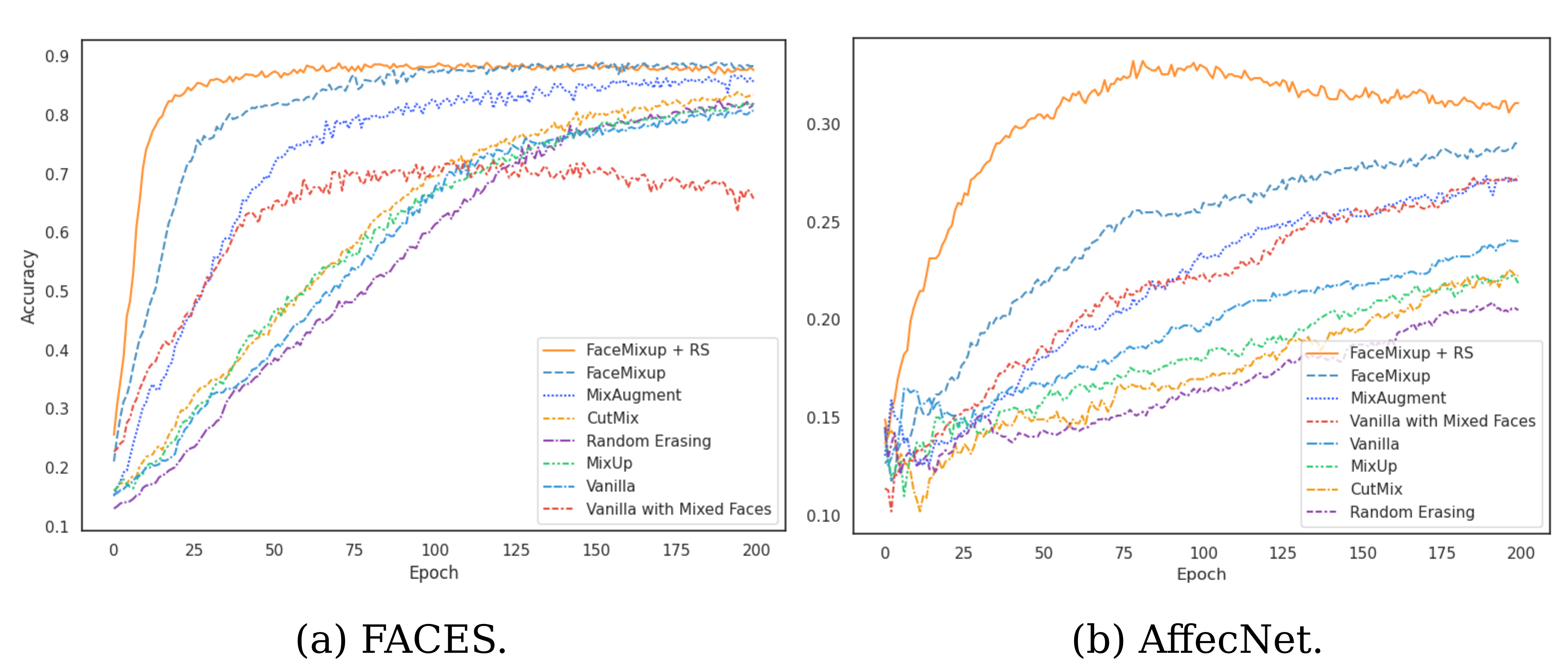}
    \caption{Learning curves of data augmentation approaches on the datasets used in this work.}
    \label{fig:acc_fer}
\end{figure*}

In this section, a more detailed analysis is performed, considering the learning curves of the data augmentation approaches compared in this work. Figure~\ref{fig:acc_fer} show the learning curves of the eight approaches compared in this work for two datasets (FACES and AffectNet).

Firstly, the superiority of \textit{FaceMixup}-based approaches (\textit{FaceMixup} and \textit{FaceMixup}$+$RS) over other baseline approaches is easily observed in both datasets. However, on the FACES dataset, these approaches have similar results from $125$ to $200$ epochs. This fact occurs because the FACES dataset is composed of facial images from the laboratory environment (controlled environment), therefore the additional component (real samples -- RS) in the loss function does not take advantage of this dataset. Unlike on the AffectNet dataset that \textit{FaceMixup}$+$RS approach achieved maximal accuracy of $33.23\%$ against $29.01\%$ achieved by \textit{FaceMixup} approach.

Another important issue to discuss is related to the role of the multiple cross-entropy functions (like Mixup approach) in the \textit{FaceMixup} loss (${\cal L}_{FaceMixup}$). This fact can be observed when \textit{FaceMixup} (dashed blue line) and Vanilla with Mixed Faces (red line) approaches are analyzed on the FACES dataset. Note that simply adding mixed facial images without adjusting the errors (backpropagation algorithm) through the combined cross-entropy functions (two classes) becomes detrimental to the EfficientNet architecture. Therefore, the mixed face images present in the training set of the Vanilla with Mixed Faces approach can cause a noisy label effect during the training process. However, on the AffectNet dataset (in-the-wild images), the addition of the MIXED FACES dataset to the training set worked as a ``nature regularization agent" for the uncontrolled images and the Vanilla with Mixed Faces approach improved it results in relation to Vanilla approach.

\begin{figure*}[ht!]
    \centering
    \includegraphics[width=0.85\linewidth]{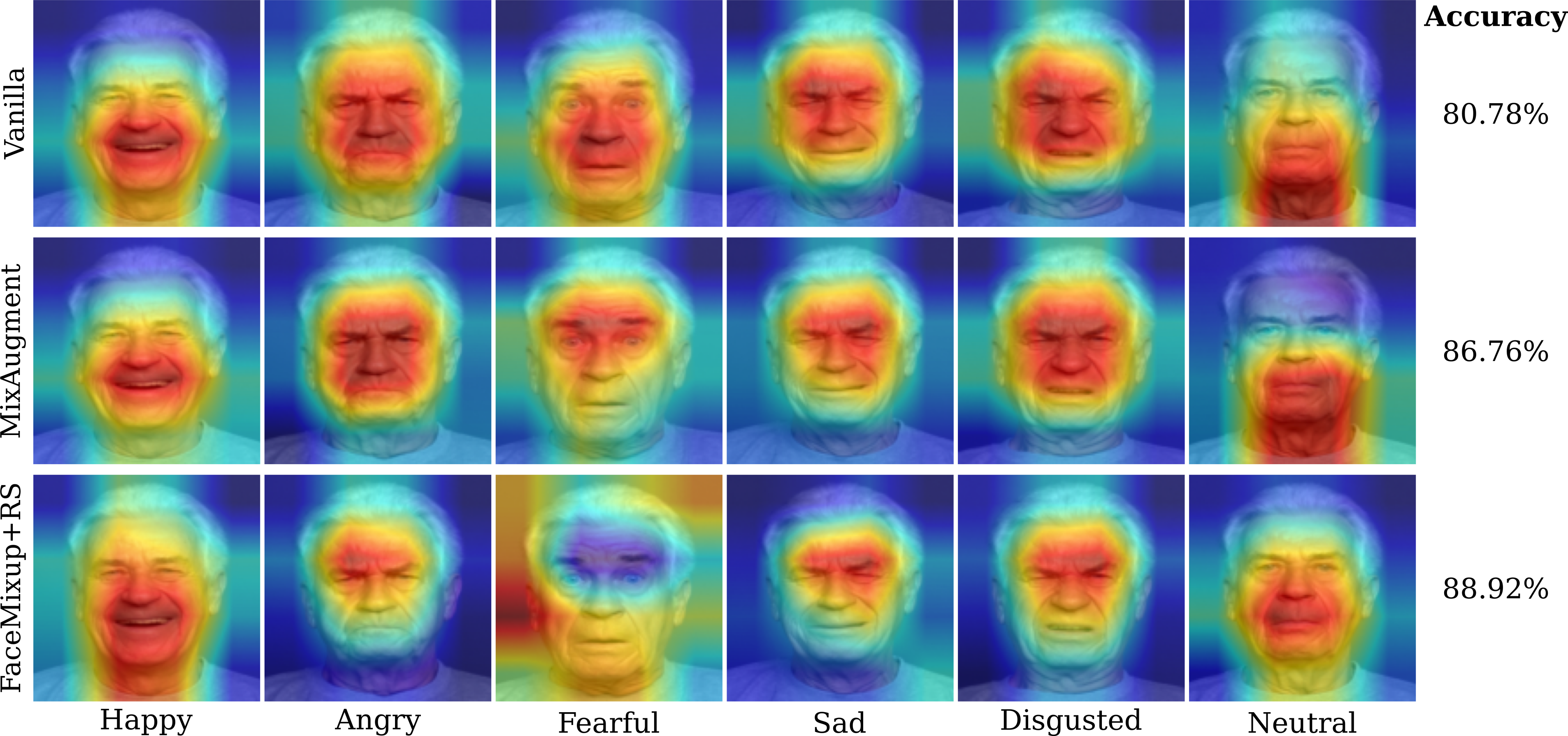}
    \caption{Qualitative comparison  among the Vanilla, MixAugment, and FaceMix$+$RS on the FACES dataset. }
    \label{fig:cam}
\end{figure*}

\begin{figure*}[ht!]
    \centering
    \includegraphics[width=1\linewidth]{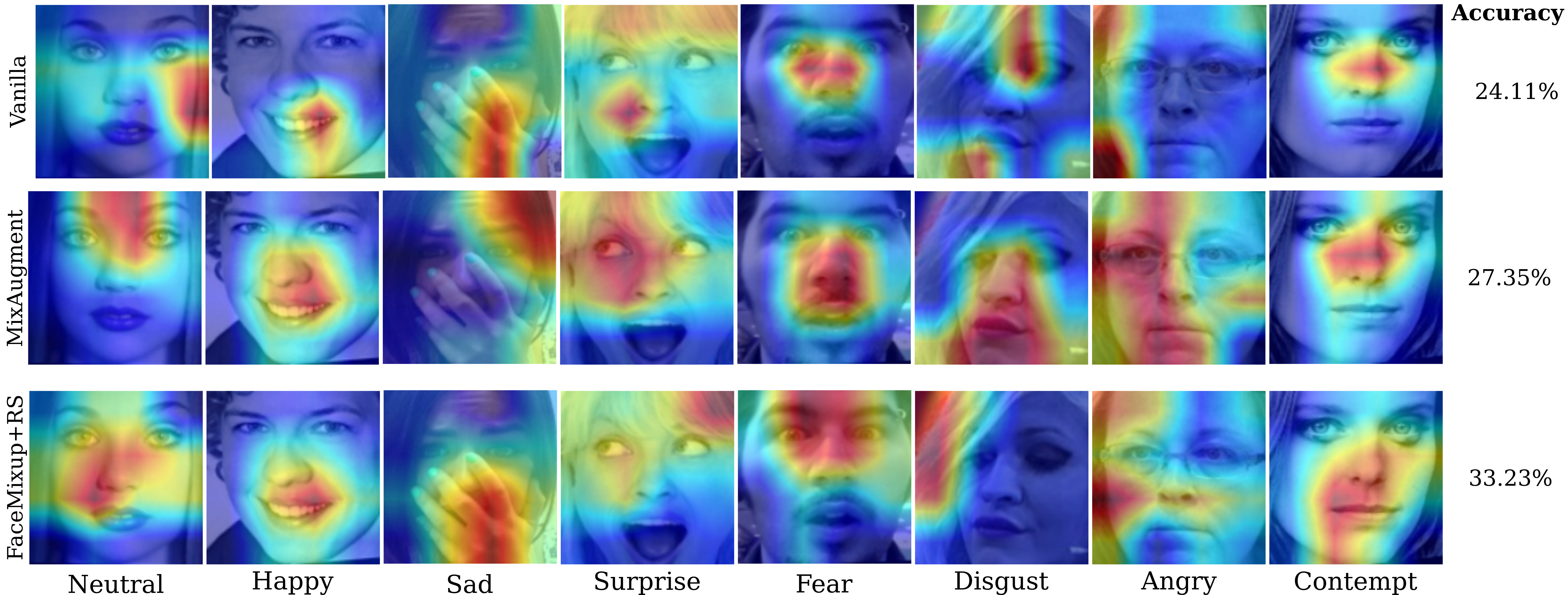}
    \caption{Qualitative comparison among the Vanilla, MixAugment, and FaceMix$+$RS on the AffectNet dataset.}
    \label{fig:cam_affectnet}
\end{figure*}

In Figures~\ref{fig:cam} and ~\ref{fig:cam_affectnet} are the qualitative results through use of class activation maps (CAM~\cite{cam_2016}) among three different approaches analyzed on the FACES and AffectNet datasets, respectively. A CAM assigns a weight/score to each element present in the analyzed image, indicating how much it contributes to the final prediction in the facial expression recognition task. The heatmap varies in range from blue (low values) to red (high values).

\section{Conclusion}
\label{sec:conclusion}

In this paper, we proposed \textit{FaceMixup} as a facial data augmentation approach for Facial Expression Recognition (FER). \textit{FaceMixup} approach is an adaptation of CutMix to replace regions between pairs of face images, but it differs from CutMix in the use of semantics information to replace the region of interest (facial components). Furthermore, \textit{FaceMixup} uses the combination of multiple cross-entropy functions from different classes in the final  loss function, as proposed by Mixup. In scenarios with uncontrolled images (in-the-wild scenario), an improved version (\textit{FaceMixup}$+$RS) was proposed, incorporating mixed and original images in the training process, as in MixAugment. In the performed experiments, six different baseline approaches (Vanilla, Vanilla with Mixed Faces, Mixup, CutMix, MixAugment, and Random Erasing) were compared with the proposed approaches (\textit{FaceMixup} and \textit{FaceMixup}$+$RS) using EfficientNet as standard CNN on two datasets (FACES and AffectNet). Furthermore, two more variations of each approach were tested using AA and Cutout techniques, totaling $16$ different experimental runs per dataset. Finally, in these experiments, two different scenarios (controlled  and in-the-wild environments) were designed to serve as validation protocols for the approaches proposed in this work. As it was possible to observe, the \textit{FaceMixup}-based approaches achieved the best accuracy results among all approaches compared in the experiments. When compared with the best baseline approach (MixAugment and Vanilla with Mixed Faces), the \textit{FaceMixup}$+$RS achieved at least relative gains of  $2.46\%$ and $7.78\%$ in the FACES and AffectNet datasets, respectively. Therefore, the \textit{FaceMixup}-based approaches indicate that they can be good alternative solutions for facial data augmentation approaches in real-world applications.

\section*{Limitations}
This section describes three limitations of the \textit{FaceMixup}-based approaches proposed in this work. (1) The \textit{FaceMixup} approach does not work on-the-fly mode, therefore it is needed to generate the transformed images in the pre-processing step and thus the current implementation is dependent on this creation of the images, increasing processing time and restricting/limiting the combinatorial power of the approach (\emph{e.g.}, $\sim$61 millions new mixed face images for the FACES dataset); (2) The \textit{FaceMixup} approach uses a fixed value for the $W$ parameter in the loss function (${\cal L}_{FaceMixup}$), which needs to be found during the experimental stage. An evolution of the approach would be an adaptive technique similar to CutMix that it is based on cut region areas; and (3) In terms of in-the-wild scenarios, a face registration/alignment procedure is necessary so that the blending of the faces is better adjusted.

\section*{Acknowledgements}
FAF is grateful to the S{\~a}o Paulo Research Foundation (FAPESP) grants 2017/25908-6 and 2018/23908-1. RFST would like to thank the Mato Grosso State Research Support Foundation –
FAPEMAT (process FAPEMAT-PRO-2022/01047) for support during development this work.

\bibliographystyle{elsarticle-num}
\bibliography{refs}

\end{document}